%% file: main.tex
\documentclass[10pt,twocolumn,letterpaper]{article}

\usepackage[pagenumbers]{iccv} 

\usepackage{times}
\usepackage{epsfig}
\usepackage{graphicx}
\usepackage{amsmath}
\usepackage{amssymb}
\usepackage{dsfont}
\usepackage{booktabs}
\usepackage{algorithmic}
\usepackage{graphicx}
\usepackage{textcomp}
\usepackage{xcolor}
\usepackage{stfloats}
\usepackage{url}
\usepackage[utf8]{inputenc}
\usepackage{fourier} 
\usepackage{array}
\usepackage{arydshln}
\usepackage{makecell}
\usepackage{multirow}
\usepackage{tabularx}
\usepackage{subcaption}
\usepackage{pifont}
\usepackage{natbib}
\usepackage{xspace}

\graphicspath{{figures/}}


\definecolor{iccvblue}{rgb}{0.21,0.49,0.74}
\usepackage[pagebackref,breaklinks,colorlinks,allcolors=iccvblue]{hyperref}

\def\paperID{3030} 
\def\confName{ICCV}
\def\confYear{2025}

\title{Learning an Ensemble Token from Task-driven Priors in Facial Analysis}

\author{Sunyong Seo, Semin Kim, Jongha Lee\\
lululab, AI R\&D center\\
Seoul, 06054, Republic of Korea\\
{\tt\small \{sy.seo, sm.kim, jongha.lee\}@lulu-lab.com}
}

\begin{document}
\maketitle
\input{sec/manuscript}
{
    \small
    \bibliographystyle{ieeenat_fullname}
    \bibliography{main}
}

\end{document}

%% file: sec/manuscript.tex
\begin{abstract}
Facial analysis exhibits task-specific feature variations. While Convolutional Neural Networks (CNNs) have enabled the fine-grained representation of spatial information, Vision Transformers (ViTs) have facilitated the representation of semantic information at the patch level. While advances in backbone architectures have improved over the past decade, combining high-fidelity models often incurs computational costs on feature representation perspective. In this work, we introduce KT-Adapter, a novel methodology for learning knowledge token which enables the integration of high-fidelity feature representation in computationally efficient manner. Specifically, we propose a robust prior unification learning method that generates a knowledge token within a self-attention mechanism, sharing the mutual information across the pre-trained encoders. This knowledge token approach offers high efficiency with negligible computational cost. Our results show improved performance across facial analysis, with statistically significant enhancements observed in the feature representations.
\end{abstract}

\section{Introduction}
\label{sec:intro}

Facial analysis exhibits task-specific feature variations; conversely, a unified feature representation derived from diverse features yields robust performance across facial analysis tasks. Deep learning neural networks represent spatial and semantic features from a single facial image. The Convolutional Neural Networks (CNNs) enable spatial information representation, and more recently, the Vision Transformers (ViTs) facilitate the semantic information representation at the patch level. Notably, the transformer's attention mechanism enables robust representation even when inputs are scattered across disparate spaces or modalities~\citep{Kirillov_2023_ICCV, Wei_2020_CVPR}. 

\begin{figure}[tbp]
  \centering
  \mbox{}
  \includegraphics[width=.8\linewidth]{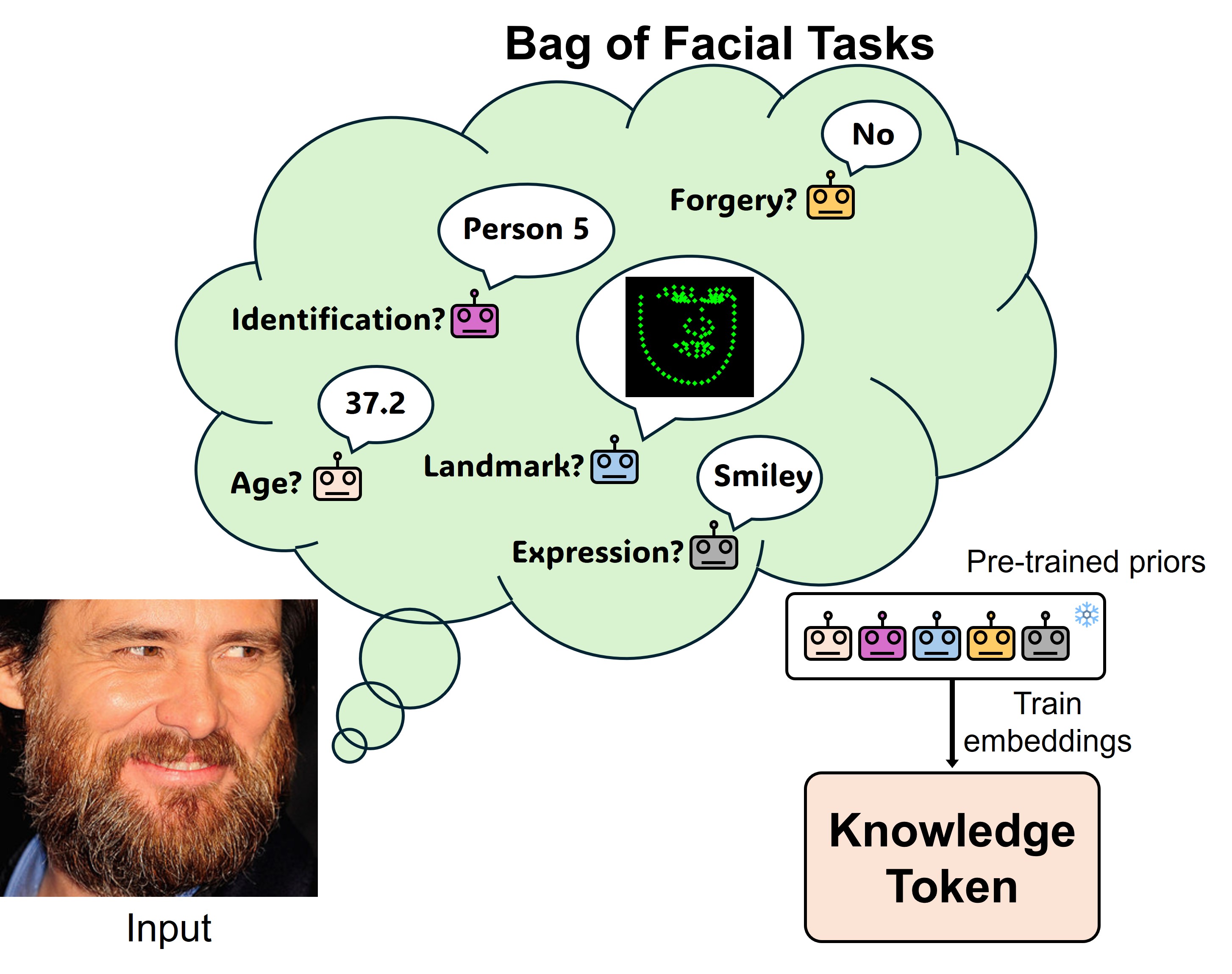}
  \caption{\label{fig:main-figure}
  The key concept of our proposed approach. Our method leverages pre-trained priors from a  frozen encoder and unifies them into a learnable token, called knowledge token.}
\end{figure}

The emergence of general models such as DINOv2~\citep{oquab2024dinov2learningrobustvisual} and SAM~\citep{kirillov2023segment} capable of handling tasks has significantly advanced interpretability within the computer vision. These general models have demonstrated significant achievements in image classification, segmentation, and detection, and as well as handling multi-tasks within a single encoder. Furthermore, linear probing of pre-trained models with unified feature representation has shown promising results~\citep{he2021maskedautoencodersscalablevision, Liang_2022_CVPR, Pham_2024_WACV, Huang_2024_CVPR}. In the areas of multi-modality, CLIP~\citep{radford2021learningtransferablevisualmodels, Shen_2023_ICCV} and attention mechanisms~\citep{vaswani2023attentionneed, dosovitskiy2021imageworth16x16words} have demonstrated promising results, confirming that unified feature representation significantly enhances model's interpretability.


As noted, the capacity of single and multi-task learning in conventional methodologies has stepped up the visual interpretability. However, there remains a paucity of research that preserves the unified feature representation or encoder's priors on single task learning during the training phase. Although task-oriented fine-tuning can be achieved through ensemble learning~\citep{Ganaie_2022, 9893798, MOHAMMED2023757}, there is still a challenge in that the computational cost escalates proportionally to both the number of tasks and the encoder model's capacity. In this research, we introduce a methodology for learning knowledge tokens by leveraging the attention mechanisms based on the task priors of pre-trained models in facial analysis tasks. The principal contributions of this study are shown in Fig.~\ref{fig:main-figure} and as follows:

\begin{itemize}
\item We propose KT-Adapter, a module for adapting and learning query priors based on self-attention mechanism, sharing the mutual information across the prior tasks.
\item The knowledge token within the KT-Adapter denotes a highly efficient method and requires negligible computational cost. Furthermore, the KT-Adapter's flexible architecture allows for straightforward adaptation and integration across domains and tasks
\item We present an efficient task-specific adaptation without training from scratch, unaffected by data limitations and hardware cost.
\end{itemize}

\section{Related works}

\begin{figure}[tbp]
  \centering
  \mbox{}
  \includegraphics[width=.85\linewidth]{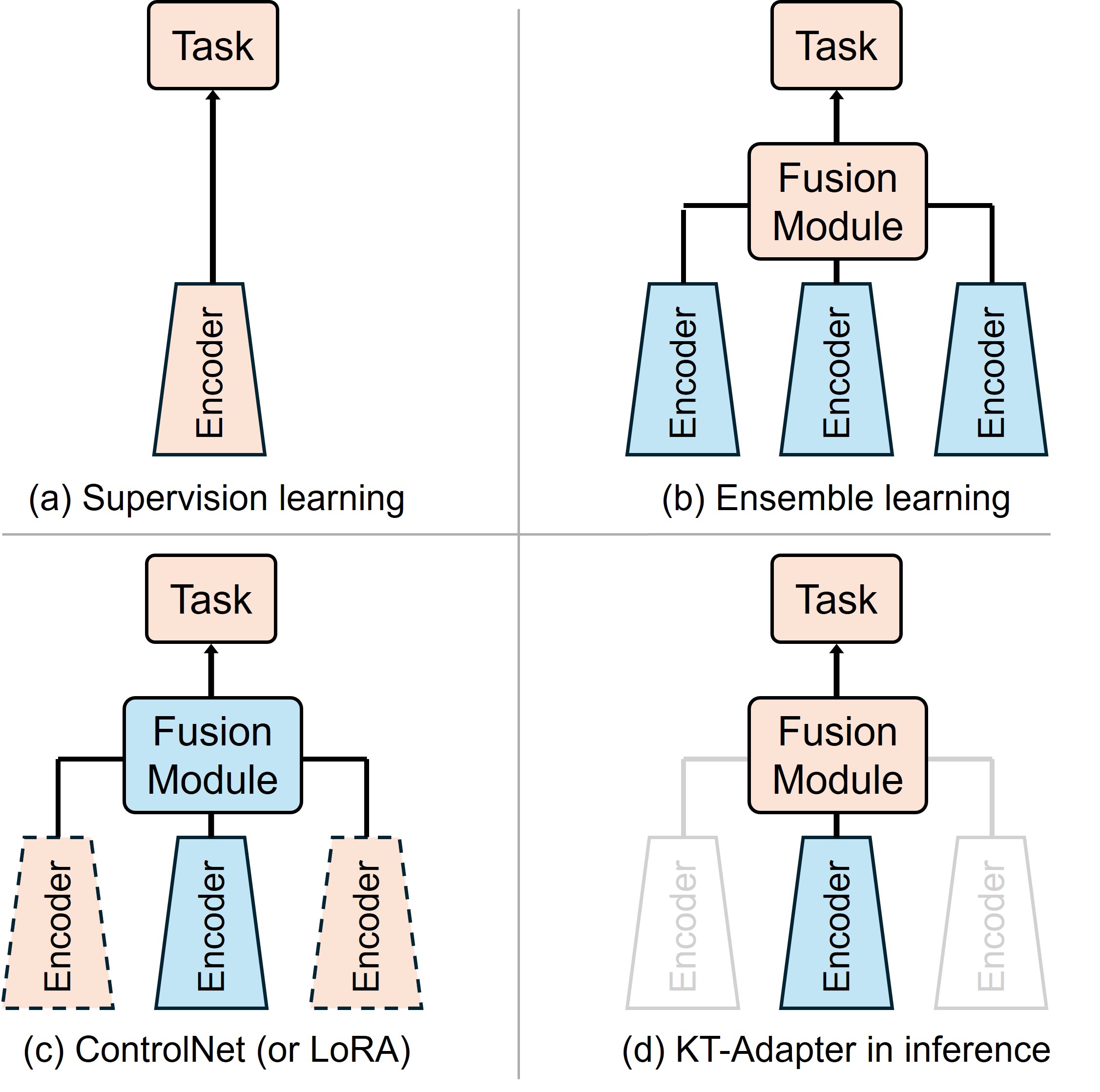}
  \caption{\label{fig:main-contribution}
  Knowledge learning typically train (red) pre-trained (blue) priors within a fusion module, while ControlNet incorporates an auxiliary lightweight (dashed-line) encoder. In contrast, KT-Adapter implicitly (grey) learns encoder priors within its fusion module, thereby reducing computational cost dramatically.}
\end{figure}

\subsection{Leveraging Pre-trained Priors}
\textbf{Ensemble learning.}
The capacity to train specialized models for individual tasks has enabled ensemble methods to achieve robust performance through the exploitation of task-specific priors. These methods range from classical techniques like bagging, boosting, and stacking, to more sophisticated fusion approaches such as majority voting and deep learning-based methodologies~\citep{Qian_2023_ICCV, Chen_2023_ICCV, 10.1007/978-3-031-73021-4_7, 10.1007/978-3-031-73209-6_18}. Nevertheless, a fundamental limitation of ensemble methodologies resides in their computational cost, which is directly proportional to both model capacity and ensemble cardinality. This computational cost can preclude their applicability in real-time facial analysis scenarios as shown in Fig.~\ref{fig:main-contribution}. Although recent studies~\citep{pmlr-v162-wortsman22a, 10444954} have explored strategies to mitigate the computational cost of model cardinality, the challenge of developing training frameworks that are simultaneously efficient, learnable, and adaptable across high feature spaces remains unresolved.


\textbf{Low rank adaptation (LoRA).}
The LoRA~\citep{hu2021loralowrankadaptationlarge} was introduced to enable parameter-efficient fine-tuning (PEFT), primarily to reduce the computational cost and hardware demands of training parameter-heavy models. Within the computer vision, LoRA has also proven valuable as a PEFT strategy, especially for diffusion models~\citep{10.1007/978-3-031-73661-2_10, 10.1007/978-3-031-72684-2_11}. Following LoRA, ControlNet-based~\citep{controlnet} adaptation of foundation models has demonstrated significant efficacy in domain and task adaptation even on multi-modality face analysis tasks~\citep{10.1007/978-3-031-72649-1_10, varanka2024localizedfinegrainedcontrolfacial}. While ControlNet has established its capacity to preserve pre-trained priors within task-specific models, the potential for ensembling encoders to achieve controllable training warrants further investigation.

\subsection{Feature Tokenization on Facial Analysis}
The attention mechanism~\citep{dosovitskiy2021imageworth16x16words} and CLIP~\citep{Ou_2023_ICCV} have become a cornerstones within the computer vision, effectively capturing both the semantic tokens and the spatial hierarchies within single and multi-modal data. In facial analysis, the HOTformer~\citep{SU2023109443} introduced a novel approach employing lightweight CNNs guided by holistic and atomic tokens. This hybrid token of HOTformer effectively integrates facial semantics and outperforms the accuracy on face recognition. The preliminary work~\citep{Zhang_2023_ICCV} proposed TokenFace, a transformer-based 3D face reconstruction model with a learnable facial component tokens encompassing shape, expressions, jaw pose, camera pose, texture, and lighting. This multi-modality approach effectively integrates the interdependencies between facial characteristics and image token embeddings derived from ViT within the single task. The POSTER++~\citep{MAO2025110951} advanced the state-of-the-art in facial emotion recognition by replacing traditional pyramid architectures with a more efficient design, achieving performance with minimal computational cost. This method leverages cross-attention to fuse landmark priors and image features. The FaceXFormer~\citep{narayan2024facexformerunifiedtransformerfacial} fuses the hierarchical representation extracted from CNN blocks. Its task-specific tokenization scheme facilitates multi-task learning within a single encoder architecture, effectively utilizing both self-attention and cross-attention mechanisms.

\begin{figure*}[tbp]
  \centering
  \mbox{}
  \includegraphics[width=.95\linewidth]{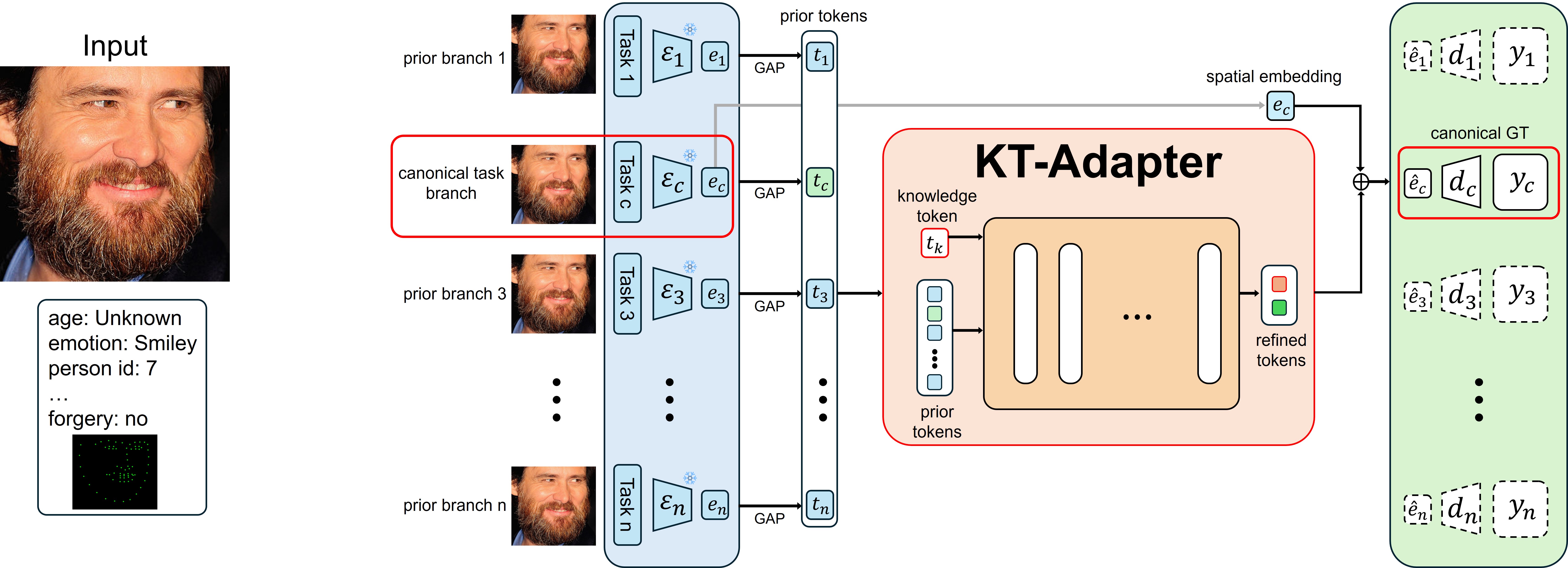}
  \caption{\label{fig:main-structure}
  Training phase of proposed method. Within the designated blue block, all encoders \(E\) are maintained as frozen layers, indicating that they are not subject to parameter updates during training. Both the KT-Adapter and the canonical branch decoder \(d_c\) represent the trainable layer within the architecture. In the green block, decoders belonging to the set \(D \setminus \{d_c\}\), which correspond to other tasks distinct from the canonical task, are excluded from the forward pass graph. Furthermore, The existence of the ground truth set \(Y \setminus \{y_c\}\), denoted as dashed line, is contingent upon the dataset, exhibiting potential variability in its presence; exceptionally, the \(y_c\) associated with the canonical branch is an indispensable element.}
\end{figure*}

While attention mechanisms have demonstrated efficacy in capturing mutual information at both coarse and fine-grained levels, and while facial analysis benefits from the flexibility of determining between CNN- and ViT-based architecture contingent on task-specific requirements, current methodologies employ a single learnable encoder for mutual information learning. This architectural characteristic can induce performance bottlenecks, particularly when constrained by encoder's capacity. So as in leveraging the pre-trained priors, ensemble learning often achieve state-of-the-art results, but their computational expense can preclude their practical application on facial analysis. Although LoRA-based architectures transform embedding representations via auxiliary layers, the use of a single, lightweight encoder prove insufficient to fully capture the overfull information encoded within each pre-trained priors. Our KT-Adapter addresses both the computational cost and the issue of prior's redundancies through the knowledge token.

\section{Methodology}
Inspired by the object token in DETR~\citep{10.1007/978-3-030-58452-8_13} and mask token in MaskFormer~\citep{NEURIPS2021_950a4152, Cheng_2022_CVPR}, we propose learnable knowledge tokens corresponding to learnable query features in preliminaries. DETR introduced an anchor-free semantic representation leveraging object tokens. Subsequently, Mask2Former utilized masked attention to incorporate both semantic and spatial information into learnable query features. Motivated by the concept of learnable query embeddings, we introduce a knowledge token designed to capture diverse embeddings derived from the priors of a pre-trained model. While conceptually similar to a module that fuses information acquired from an ensemble, this knowledge token is specifically tailored to the canonical task, selectively attending to and learning pertinent semantic information.

\subsection{Acquiring Prior Tokens}
In the preliminary stage, a set of \(N\) pre-trained encoders \(E = \{\varepsilon_1, \varepsilon_2, ..., \varepsilon_{n-1}, \varepsilon_c\}\) corresponding to the expertise of each task is prepared. These encoders could be employed either CNN or ViT architectures, or even a hybrid thereof dictated by task-specific priors. Subsequent to the preparation of \(E\), a canonical task branch is selected for task-specific training. As illustrated in Fig.~\ref{fig:main-structure}, face landmark detection serves as the canonical task in the presented example. During training stage, supervision learning is exclusively derived from the ground truth \(Y = \{y_1, y_2, ..., y_{n-1}, y_c\}\) corresponding to the designated canonical task; ground truth of the \(N-1\) remaining tasks \(Y \setminus \{y_c\}\) is explicitly excluded from the training process.

Within the prior branches, \(\varepsilon_i\) processes an identical input image, generating a corresponding spatial embedding \(e_i\) from prior branch and \(e_c\) from canonical task branch. Subsequently, global average pooling (GAP) is applied to \(e_i\) to derive a prior tokens \(T = \{t_1, t_2, ..., t_{n-1}, t_c\}\). These prior tokens represent disparate embeddings, each encapsulating distinct characteristics that diverge from those of the canonical task. The prior token derived from the canonical task branch is denoted as \(t_c\). To accommodate variations in prior token dimensionality arising from differing encoder capacities, zero-padding is employed to mask any remaining space up to the uniform size. Zero-padding enforces dimensional uniformity by masking unused spatial positions with zeros, thereby obviating superfluous linear transform computations.

\begin{figure*}[tbp]
  \centering
  \mbox{}
  \includegraphics[width=.95\linewidth]{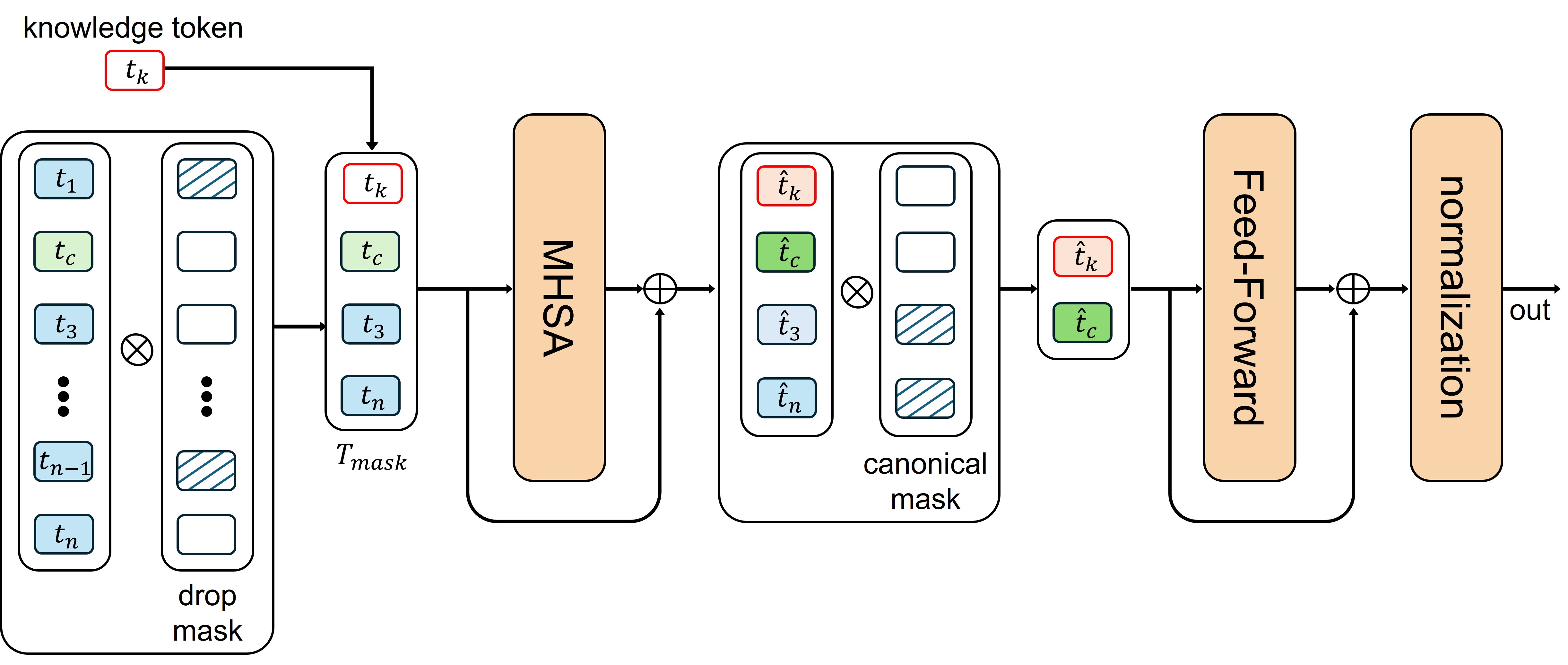}
  \caption{\label{fig:KT-Adapter}
  Detailed structure of KT-Adapter. The KT-Adapter is trained on a restricted information \(T_{\text{mask}}\) to enrich the \(e_c\) in Fig.~\ref{fig:main-structure}. During the training phase, the MHSA layer processes \(T_{\text{mask}}\), which has been subjected to subsampling via the drop mask. During inference, the MHSA layer only processes \(\{t_k, t_c\}\), and the canonical mask is omitted.}
\end{figure*}

\subsection{KT-Adapter}
We introduce the KT-Adapter architecture to enable the learning of knowledge tokens. KT-Adapter functions to adapt a knowledge token, drawing inspiration from class token within attention mechanisms and object tokens in DETR~\citep{10.1007/978-3-030-58452-8_13}, to a designated canonical task through the utilization of prior tokens. Leveraging a self-attention mechanism, KT-Adapter facilitates the learning of both knowledge token \(t_k\) in KT-Adapter and canonical prior \(t_c\) derived from canonical encoder \(e_c\). By focusing on the canonical task, it effectively reduces redundancy among the remaining prior tokens, which represent task-optimized embeddings from pre-trained tasks, thereby generating representations \(\hat{t}_k\) and \(\hat{t}_c\). Distinct from ensemble and ControlNet architectures, the KT-Adapter architecture eliminates the requirement for supplementary priors during inference, thereby enabling computationally efficient inference based solely on the learned knowledge token.

\subsection{Drop mask and Canonical mask}
In Fig.~\ref{fig:KT-Adapter}, we initialize the \(t_k\), a learnable query feature, and subsequently concatenate it with the \(T\). Drawing inspiration from MaskFormer~\citep{NEURIPS2021_950a4152} and dropout~\citep{10.5555/2627435.2670313}, random subsampling within the feature representation not only mitigates overfitting but also rectifies global representations and potential misalignment in the token's forward pass between training and inference phases, specifically within KT-Adapter. We introduce a stochastic drop mask \(m\) denoted as:

\begin{equation} \label{eq:drop-mask1}
\ m_i =
\begin{cases}
1, & \text{if } \ x_i > \theta \ \text{, or} \  i = c \\ 
0, & \text{otherwise}
\end{cases}
\end{equation}

\begin{equation} \label{eq:drop-mask2}
T_{mask} = \{t_i \in T \mid m_i = 1 \},
\end{equation}
which randomly subsamples the \(T\) over threshold \(\theta\). The \(T\) are subjected to drop mask and consequently excluded from further processing. During the construction of the \(m_i\), the \(t_c\) is invariably passed through the subsampling, ensuring inference consistency and conferring a higher magnitude relative to other prior tokens, mirroring the treatment of the knowledge token itself. 

We employed multi-head self-attention (MHSA) for learning knowledge tokens. The subsampled token set \(T_{{masked}}\) in Eq.~\ref{eq:drop-mask2} is presented as input to the conventional attention mechanism:

\begin{equation} \label{eq:attention1}
\text{Attention}(Q, K, V) =\text{SoftMax}\left(\frac{QK^T}{\sqrt{d}}\right) \cdot V
\end{equation}

In application of self-attention, positional embeddings are applied to \(T_{\text{masked}}\) based on the primitive task position \(i\). The \(\hat{T}\) generated by the attention layer undergoes subsampling to \({\hat{t}_k, \hat{t}_c}\) via a canonical mask. Significantly analogous to the stochastic drop mask, the canonical mask serves to mitigate potential inconsistencies between the training and inference phases. In subsequent, a feed-forward network and normalization layer facilitate a seamless transition of the representations \({\hat{t}_k, \hat{t}_c}\) to the decoder pathway.

\subsection{Decoder and Inference}
\label{sec:decoder-and-inference}
The \(\hat{t}_k\) and \(\hat{t}_c\) derived from the KT-Adapter are subjected to element-wise addition with the spatial embedding information \(e_i\). The feature embedding \(\hat{e}_c\) is subsequently provided as input to the canonical decoder \(d_c\). While a set of \(N\) decoders \(D = \{d_1, d_2, ..., d_{n-1}, d_c\}\) is potentially learnable for other tasks, only the task-specific decoder \(d_c\) undergoes fine-tuning to accommodate \(\hat{e}_c\). This canonical, or selective, fine-tuning strategy restricts the trainable parameters to the KT-Adapter and the canonical decoder, thereby significantly reducing the computational resources required for both training and inference. Each decoder comprises a light multi-layer perceptron architecture tailored for classification and regression tasks. For the loss function, WingLoss~\citep{Feng_2018_CVPR} was employed for regression tasks, including age estimation and landmark detection, and cross-entropy loss was employed for classification tasks, including face emotion recognition and face recognition.

\begin{figure}[tbp]
  \centering
  \mbox{}
  \includegraphics[width=.9\linewidth]{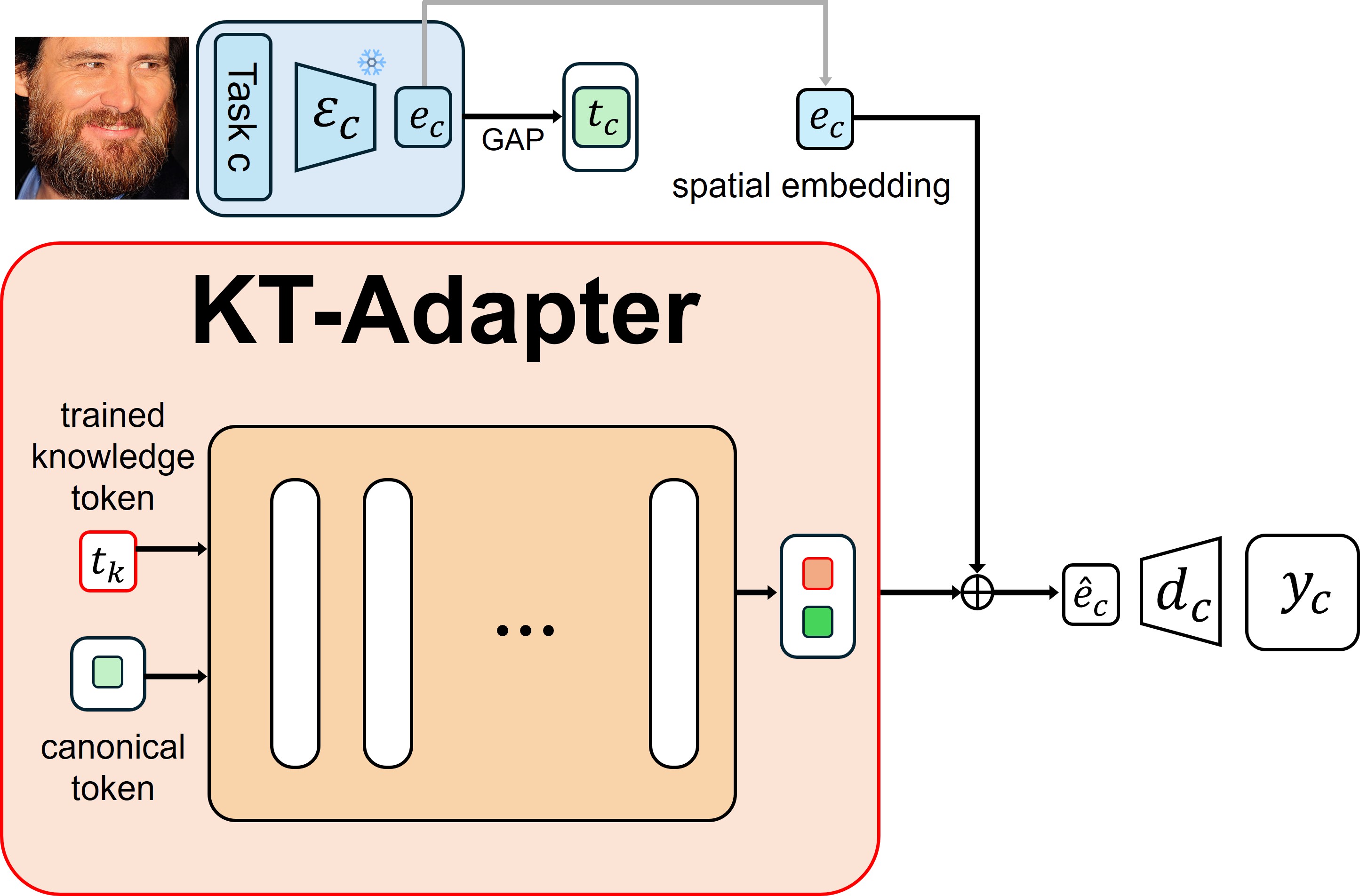}
  \caption{\label{fig:main-inference}
  Inference phase of decoupled encoder. The architectural parameters of all layers are maintained consistently with those employed during the training phase. Only the difference lies in the input provided to the KT-Adapter module, which is exclusively the canonical token.}
\end{figure}

During inference, only \(t_c\) derived from \(e_c\) are presented as input to the KT-Adapter as shown in Fig.~\ref{fig:main-inference}. While the information density is lower than training, \(t_k\) retains a sophisticated and entangled representation of the priors acquired during the training phase. The KT-Adapter's training on "thinned" information, facilitated by the drop mask, ensures that the inference process exhibits a high degree of fidelity to the training phase. To facilitate the application of positional embeddings to \(T_{\text{masked}}\), the task index corresponding to each encoder is preserved throughout the training phase. This preservation guarantees the consistent forward pass instead of positional embeddings to corresponding token's locations, even in scenarios where elements from \(T \setminus {t_c}\) are absent in inference phase. Canonical masking operations are omitted during inference as it identically equals to \(\{\hat{t}_k, \hat{t}_c\}\). In consequence, the exclusion of other \(e_i\) results in a substantial reduction in computational overhead compared to knowledge learning, with the computational savings scaling linearly with the number of encoders.

\section{Experiments}

\begin{table*}[htp]
    \centering%
    \tiny
    \renewcommand{\arraystretch}{0.85}
    \resizebox{1.0 \textwidth}{!}{
    \begin{tabular}{ c | l l | l l | l l | l l}
        \Xhline{2\arrayrulewidth}
         & \multicolumn{2}{c}{\textbf{WFLW}} & \multicolumn{2}{c}{\textbf{UTK-face}} & \multicolumn{2}{c}{\textbf{Affect-Net-HQ}} & \multicolumn{2}{c}{\textbf{CFP-dataset}} \\
        \hline
        \textbf{Structure} & NME \(\downarrow\) & MAE \(\downarrow\) & CS@5 \(\uparrow\) & MAE \(\downarrow\) & Accuracy \(\uparrow\) & F1-score \(\uparrow\) & Accuracy \(\uparrow\) & F1-score \(\uparrow\) \\
        \hline
        \hline
        M (baseline) & 5.016 & 3.104 & 66.79 & 4.568 & 72.61 & 70.23 & 88.13 & 85.04 \\
        Model Ensemble (concat) & 5.181 \color[HTML]{D62728}{(+0.165)} & 3.205 \color[HTML]{D62728}{(+0.100)} & 67.24 \color[HTML]{1F77B4}{(+0.44\%)} & 4.556 \color[HTML]{1F77B4}{(-0.012)} & \textbf{73.59 \color[HTML]{1F77B4}{(+0.98\%)}} & \textbf{71.16 \color[HTML]{1F77B4}{(+0.93\%)}} & 84.93 \color[HTML]{D62728}{(-3.20\%)} & 80.91 \color[HTML]{D62728}{(-4.13\%)} \\
        Model Soup (Uniform) & 7.864 \color[HTML]{D62728}{(+2.848)} & 4.738 \color[HTML]{D62728}{(+1.632)} & 50.59 \color[HTML]{D62728}{(-16.2\%)} & 6.388 \color[HTML]{D62728}{(+1.820)} & 69.86 \color[HTML]{D62728}{(-2.75\%)} & 67.14 \color[HTML]{D62728}{(-3.09\%)} & 86.00 \color[HTML]{D62728}{(-2.13\%)} & 81.64 \color[HTML]{D62728}{(-3.40\%)} \\
        \textbf{KT-Adapter (Ours)} & \textbf{4.976 \color[HTML]{1F77B4}{(-0.040)}} & \textbf{3.080 \color[HTML]{1F77B4}{(-0.024)}} & \textbf{67.29 \color[HTML]{1F77B4}{(+0.50\%)}} & \textbf{4.548 \color[HTML]{1F77B4}{(-0.020)}} & 73.09 \color[HTML]{1F77B4}{(+0.48\%)} & 70.63 \color[HTML]{1F77B4}{(+0.40\%)} & \textbf{89.60 \color[HTML]{1F77B4}{(+1.47\%)}} & \textbf{86.26 \color[HTML]{1F77B4}{(+1.22\%)}} \\
 
        \hline
        C (baseline) & 4.398 & 2.735 & 68.32 & 4.316 & 74.67 & 72.33 & 93.20 & 91.36 \\
        Model Ensemble (concat) & 4.459 \color[HTML]{D62728}{(+0.061)} & 2.775 \color[HTML]{D62728}{(+0.040)} & 68.60 \color[HTML]{1F77B4}{(+0.27\%)} & 4.285 \color[HTML]{1F77B4}{(-0.031)} & 74.41 \color[HTML]{D62728}{(-0.26\%)} & 72.18 \color[HTML]{D62728}{(-0.15\%)} & 92.26 \color[HTML]{D62728}{(-0.94\%)} & 89.45 \color[HTML]{D62728}{(-1.91\%)} \\
        Model Soup (Uniform) & 7.601 \color[HTML]{D62728}{(+3.203)} & 4.663 \color[HTML]{D62728}{(+1.928)} & 56.72 \color[HTML]{D62728}{(-11.6\%)} & 5.688 \color[HTML]{D62728}{(+1.372)} & 67.63 \color[HTML]{D62728}{(-7.04\%)} & 64.56 \color[HTML]{D62728}{(-7.77\%)} & 91.73 \color[HTML]{D62728}{(-1.47\%)} & 88.97 \color[HTML]{D62728}{(-2.39\%)} \\
        \textbf{KT-Adapter (Ours)} & \textbf{4.350 \color[HTML]{1F77B4}{(-0.048)}} & \textbf{2.708 \color[HTML]{1F77B4}{(-0.027)}} & \textbf{69.30 \color[HTML]{1F77B4}{(+0.98\%)}} & \textbf{4.277 \color[HTML]{1F77B4}{(-0.039)}} & \textbf{74.72 \color[HTML]{1F77B4}{(+0.05\%)}} & \textbf{72.45 \color[HTML]{1F77B4}{(+0.12\%)}} & \textbf{94.66 \color[HTML]{1F77B4}{(+1.46\%)}} & \textbf{92.71 \color[HTML]{1F77B4}{(+1.35\%)}} \\
        
        \hline
        S (baseline) & 4.543 & 2.829 & 67.49 & 4.412 & 73.90 & 71.76 & 89.20 & 85.84 \\
        Model Ensemble (concat) & 4.691 \color[HTML]{D62728}{(+0.148)} & 2.912 \color[HTML]{D62728}{(+0.082)} & 67.55 \color[HTML]{1F77B4}{(+0.06\%)} & 4.410 \color[HTML]{1F77B4}{(-0.002)} & 75.22 \color[HTML]{D62728}{(+1.32\%)} & 72.87 \color[HTML]{D62728}{(+1.11\%)} & 89.60 \color[HTML]{1F77B4}{(+0.40\%)} & 85.73 \color[HTML]{1F77B4}{(+0.11\%)} \\
        Model Soup (Uniform) & 7.428 \color[HTML]{D62728}{(+2.885)} & 4.529 \color[HTML]{D62728}{(+1.700)} & 57.58 \color[HTML]{D62728}{(-9.91\%)} & 5.477 \color[HTML]{D62728}{(+1.065} & 70.65 \color[HTML]{D62728}{(-3.25\%)} & 67.87 \color[HTML]{D62728}{(-3.89\%)} & 89.46 \color[HTML]{1F77B4}{(+0.26\%)} & 85.19 \color[HTML]{D62728}{(-0.65\%)} \\
        \textbf{KT-Adapter (Ours)} & \textbf{4.512 \color[HTML]{1F77B4}{(-0.031)}} & \textbf{2.804 \color[HTML]{1F77B4}{(-0.025)}} & \textbf{67.60 \color[HTML]{1F77B4}{(+0.11\%)}} & \textbf{4.400 \color[HTML]{1F77B4}{(-0.012)}} & \textbf{75.20 \color[HTML]{1F77B4}{(+1.30\%)}} & \textbf{72.77 \color[HTML]{1F77B4}{(+1.01\%)}} & \textbf{89.73 \color[HTML]{1F77B4}{(+0.53\%)}} & \textbf{86.44 \color[HTML]{1F77B4}{(+0.60\%)}} \\
        
        \Xhline{2\arrayrulewidth}
    \end{tabular}
    }
    \caption{\label{tab:comparative-models}  
        Comparisons to other methods. While Model Ensemble entails a time complexity of \(O(N)\) both Model Soup and our proposed approach achieve a substantially lower complexity of approximately \(O(1)\).
    }
\end{table*}

\subsection{Implementation Details}
The experimental evaluations were conducted on 2 NVIDIA Tesla H100 GPUs. For the backbone network architecture, MobileOne~\citep{Vasu_2023_CVPR} was selected to assess real-time performance on resource-constrained platforms, such as embedded systems and handheld devices, while ConvNeXt-b (CNN-based)~\citep{Woo_2023_CVPR} and Swin-b (ViT-based)~\citep{Liu_2021_ICCV} were employed to evaluate performance on desktop-level hardware. All input images were resized to 224\(\times\)224 pixels, with facial regions cropped using a YOLO-based~\citep{Redmon_2016_CVPR} face detector. No facial alignment procedures were applied; however, task-specific pre-processing pipelines were implemented to ensure robust generalization by tasks. Specifically, for age estimation and face recognition tasks, the facial region was cropped to encompass both the facial area and hair, whereas for face landmark detection and face emotion recognition, a more tightly cropped region was adapted. A uniform batch size of 32 was maintained across all tasks. Given the head-tuning paradigm employed, which is analogous to linear probing, a learning rate of 1e-5 was adopted in trainable layers with warm-up cosine scheduler. In terms of computational cost, the KT-Adapter exhibited 21.02M FLOPs during the inference phase, with a token dimension size of 1024. During the training phase, the computational cost increases proportionally with the number of tasks.

\subsection{Datasets}
We utilized the UTK-face~\citep{zhang2017ageprogressionregressionconditionaladversarial} for age estimation, the AffectNet-HQ~\citep{8013713} for face emotion recognition, the WFLW~\citep{wayne2018lab} for face landmark detection, and the CFP-dataset~\citep{7477558} for face recognition. The experimental evaluation utilized four distinct facial analysis datasets: UTK-face, a large-scale repository containing over 20,000 diverse facial images annotated for age, gender, and ethnicity, exhibiting substantial intra-dataset variability; AffectNet, a comprehensive dataset of over one million internet-sourced facial images, with approximately 440,000 manually annotated for seven discrete facial expressions and valence-arousal intensity, designed to facilitate research in automated facial expression recognition; WFLW, a dataset comprising 10,000 in-the-wild facial images with 98 manually annotated landmarks and rich attribute annotations, enabling comprehensive algorithm analysis under significant variations in expression, pose, and occlusion within a unified evaluation framework; and CFP-dataset, consisting of 500 individuals, balanced for gender and aiming for racial diversity, with 10 frontal and 4 profile facial images per individual, rigorously filtered for pose accuracy using human annotation and automated detection. For the scope of this paper, only frontal facial images were considered.

\begin{figure}[tp]
    \centering
    \centering
    \begin{subfigure}{0.47\linewidth}
        \includegraphics[width=1.\linewidth]{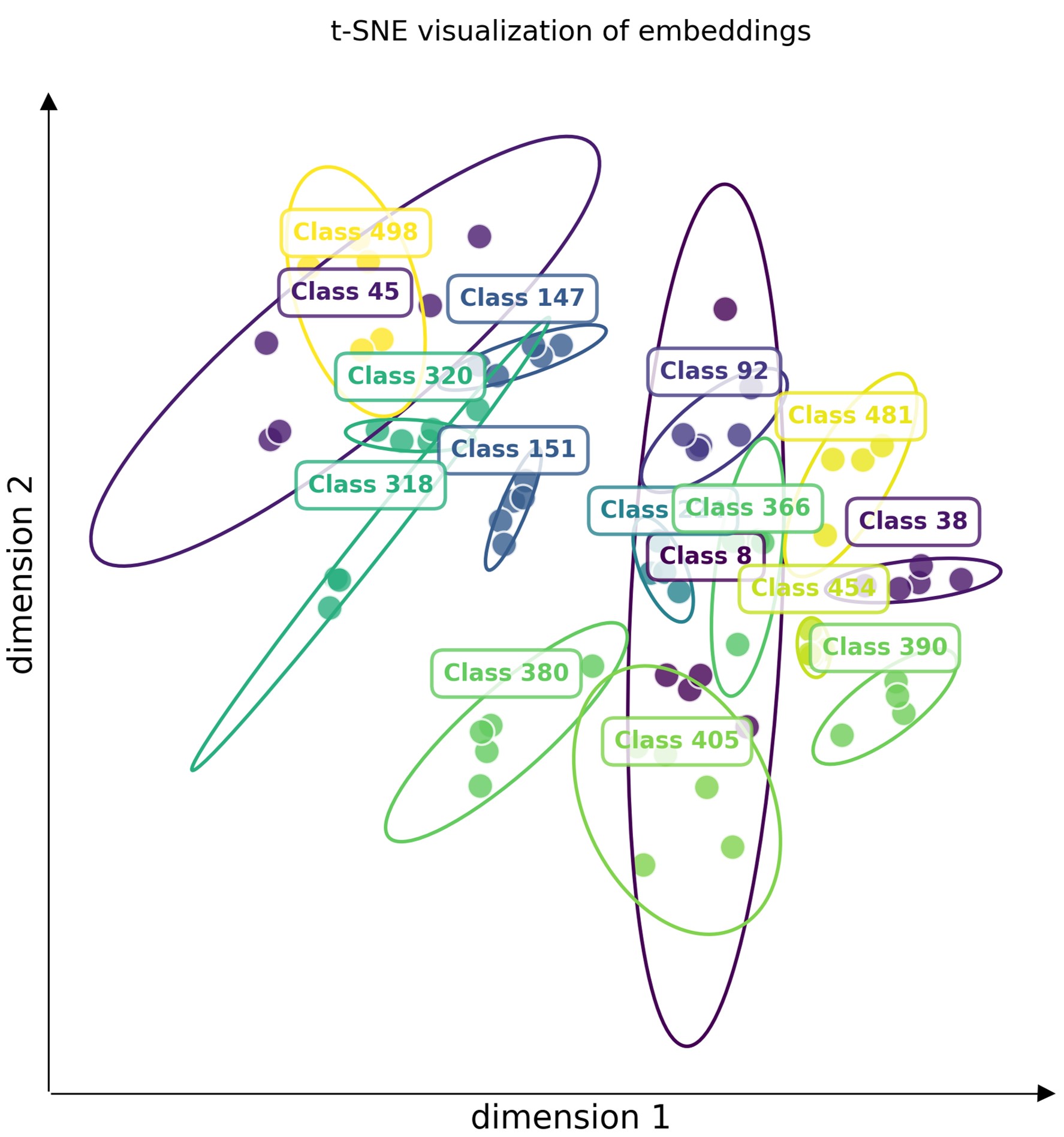}
        \caption{M (baseline)}
        \label{fig:embedding-fr-a}
    \end{subfigure}
    \hfill
    \begin{subfigure}{0.47\linewidth}
        \includegraphics[width=1.\linewidth]{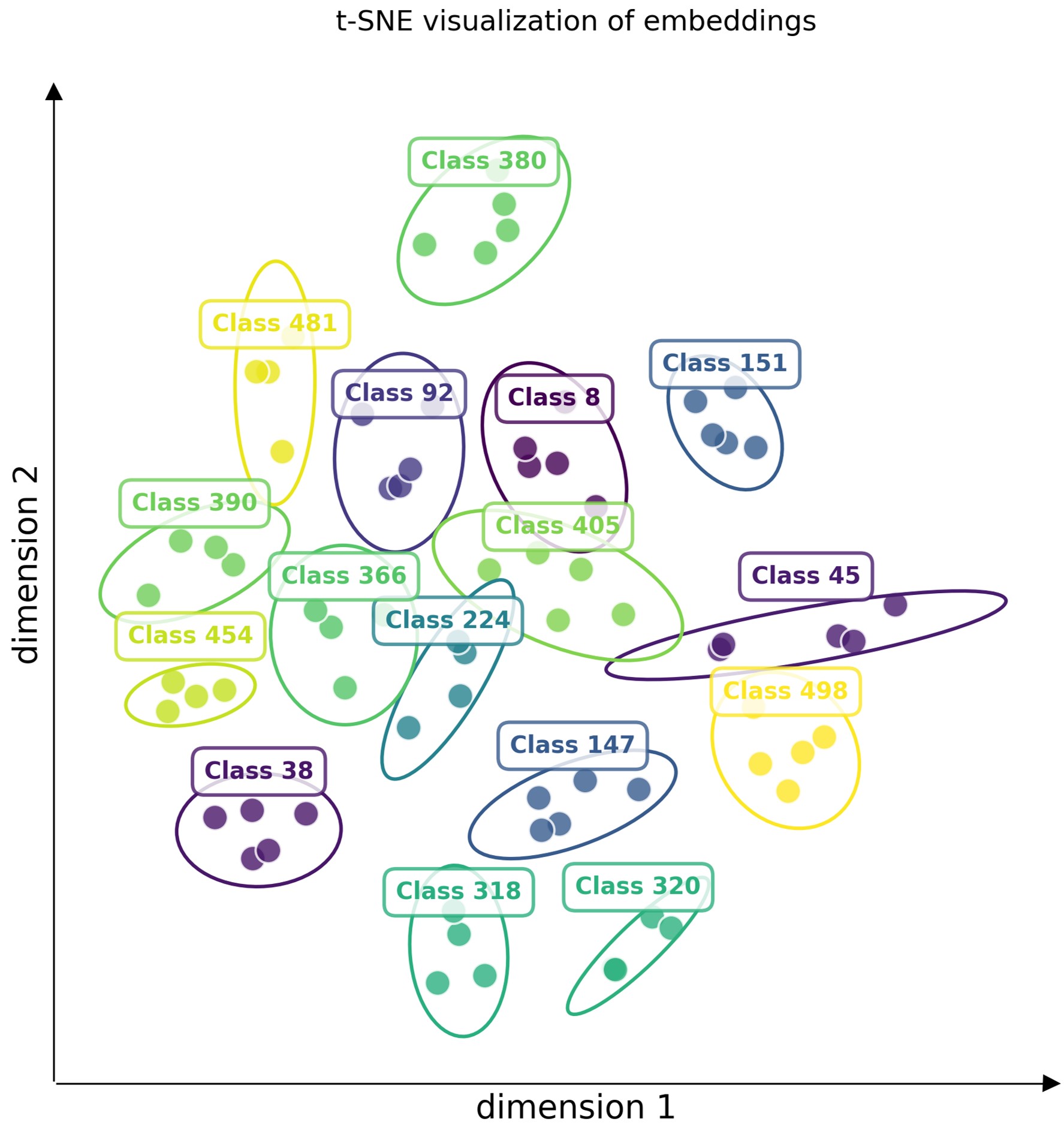}
        \caption{C to M (ours)}
        \label{fig:embedding-fr-b}
    \end{subfigure}
    
    \caption{
        Visualization of embeddings in CFP-dataset. Each color delineates a unique category within individual person identity. The elliptical contours visually depict the intra-class variance for each respective class. As corroborated by Tab.~\ref{tab:CFP-dataset}, the intra-class variance and the distinction of classes is significantly improved in the Fig.~\ref{fig:embedding-fr-b} compared to the Fig.~\ref{fig:embedding-fr-a}}
    \label{fig:embedding-fr}
\end{figure}

\begin{figure}[tp]
    \centering
    \begin{subfigure}{0.42\linewidth}
        \includegraphics[width=1.\linewidth]{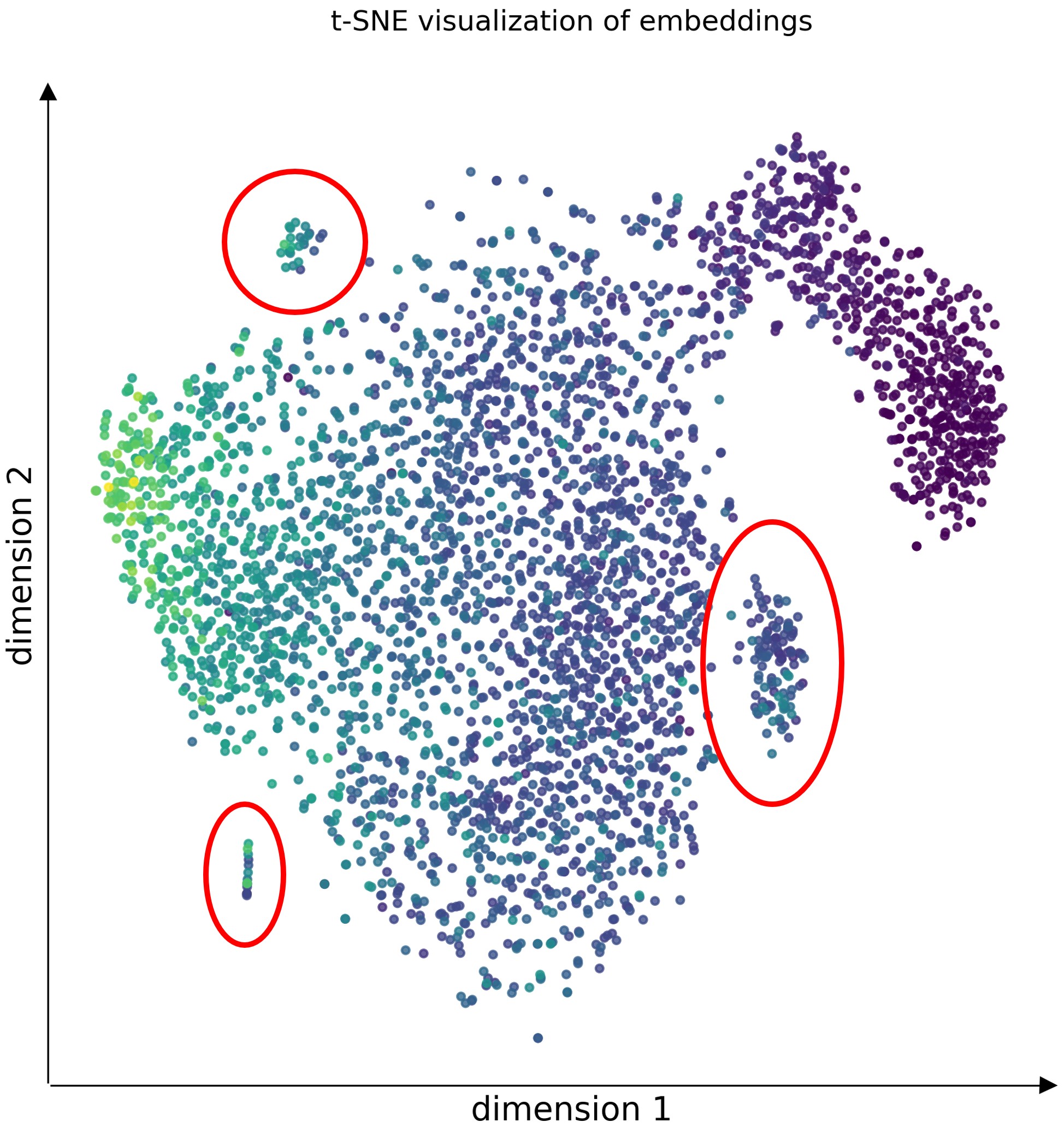}
        \caption{M (baseline)}
        \label{fig:embedding-age-a}
    \end{subfigure}
    \hfill
    \begin{subfigure}{0.5\linewidth}
        \includegraphics[width=1.\linewidth]{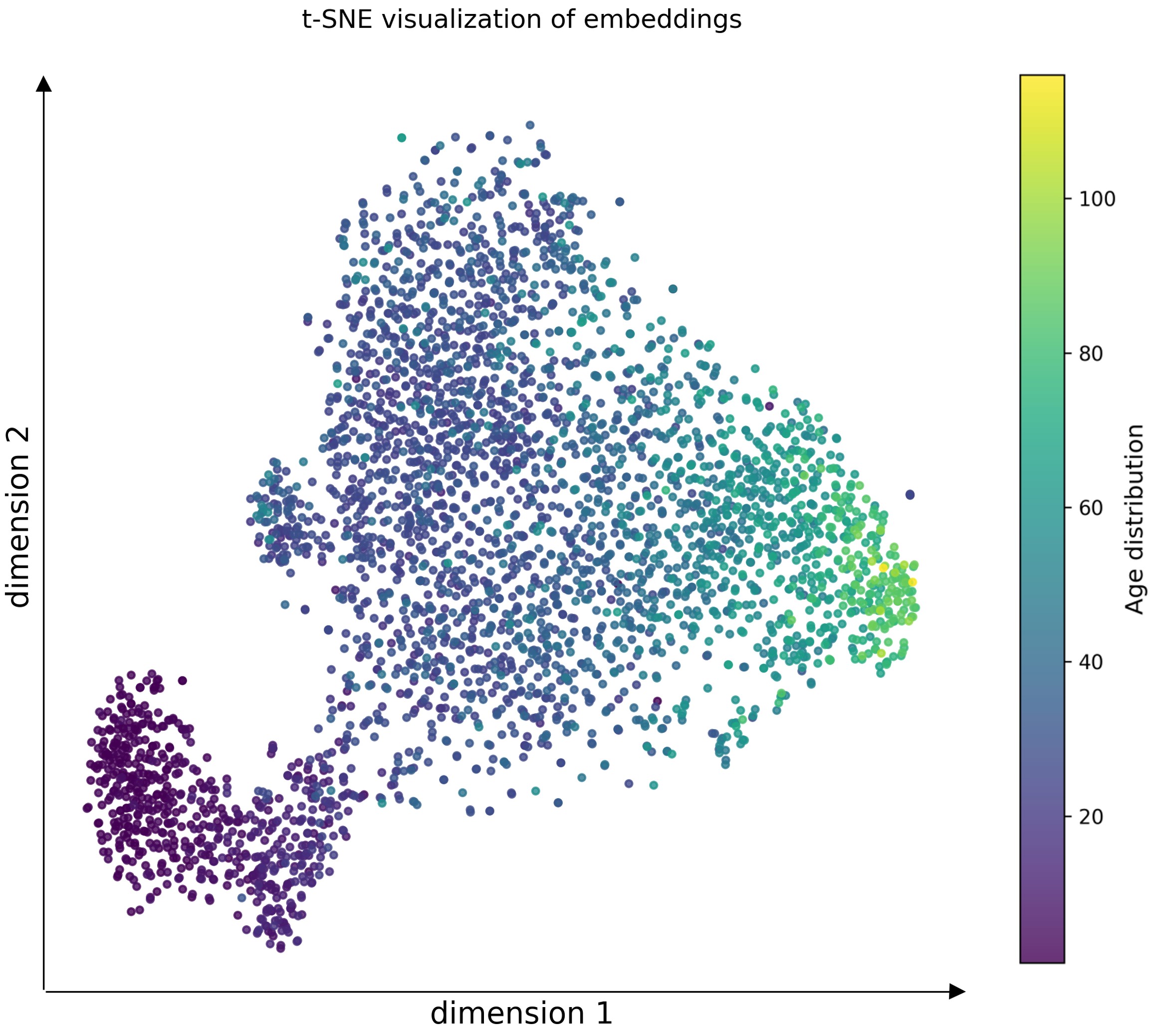}
        \caption{C to M (ours)}
        \label{fig:embedding-age-b}
    \end{subfigure}
    \caption{
        Visualization of embeddings (token) in UTK-face dataset. The color of each data point corresponds to the age represented in the right color-bar. Notably, data points that were distinct within the red ellipse in Fig.~\ref{fig:embedding-age-a} are not observed in the "C to M" data points in Fig.~\ref{fig:embedding-age-b}. This observation suggests that the KT-Adapter not only assimilate the corner-case scenarios but also represent the feature clustering with adaptable transforms.
    }
    \label{fig:embedding-age}
\end{figure}

\subsection{Pre-training the Prior Models}
The experimental design was formulated to assess the variability in feature representations arising from disparities in model architecture and capacity. Accordingly, MobileOne-s0 (M), ConvNeXt V2-base (C), and Swin Transformer-base (S) were chosen as baselines, representing lightweight CNN, heavyweight CNN, and hierarchical ViT architectures, respectively. Each model underwent pre-training on four prior tasks, utilizing a simple MLP decoder head. In terms of computational cost, considering that MobileOne-s0 operates at about 275M FLOPs, KT-Adapter's 21.02M FLOPs introduces a marginal computational burden. In comparison to other computationally demanding models, this overhead is effectively inconsequential.

For illustrative examples as experimental table, "\textbf{M (baseline)}" denotes the architectural structure pre-trained on MobileOne, while "\textbf{C to M}" denotes a configuration wherein ConvNeXt serves as the pre-trained model for the \(N-1\) prior branches, and MobileOne is designated as the canonical branch for knowledge token learning. For visualizing the embeddings, we employed the t-SNE~\citep{van2008visualizing} to reduce the N-dimensional embeddings to two-dimensional embeddings in Fig.~\ref{fig:embedding-age} and Fig.~\ref{fig:embedding-fr}.

\subsection{Evaluations on target task}

As shown in Tab.~\ref{tab:comparative-models}, in order to ensure a fair comparison, all comparative models were trained with encoders frozen while only the decoders were fine-tuned. All encoder architectures were kept identical to that of the baseline. The results indicate that both Model Ensemble~\citep{lakshminarayanan2017deepensembles} and Model Soup~\citep{pmlr-v162-wortsman22a} exhibit instability when the pre-processing procedures and target tasks are not aligned, especially on landmark detection and age estimation. In contrast, our method demonstrates remarkable robustness, maintaining strong performance even when trained across datasets and tasks.

\begin{table}[tp]
    \centering%
    \tiny
    \renewcommand{\arraystretch}{0.85}
    \resizebox{0.35\textwidth}{!}{%
    \begin{tabular}{ l | l  l }%
        \Xhline{2\arrayrulewidth}
        \textbf{Structure} & NME \(\downarrow\) & MAE \(\downarrow\) \\
        \hline
        \hline
        
        M (baseline) & 5.016 & 3.104 \\
        M to M & 4.984 \color[HTML]{1F77B4}{(-0.032)} & 3.084 \color[HTML]{1F77B4}{(-0.020)} \\
        C to M & 4.976 \color[HTML]{1F77B4}{(-0.040)} & 3.080 \color[HTML]{1F77B4}{(-0.024)} \\
        S to M & 5.020 \color[HTML]{D62728}{(+0.004)} & 3.102 \color[HTML]{1F77B4}{(-0.002)} \\
        \hline
        C (baseline) & 4.398 & 2.735 \\
        M to C & 4.360 \color[HTML]{1F77B4}{(-0.038)} & 2.712 \color[HTML]{1F77B4}{(-0.022)} \\
        C to C & 4.350 \color[HTML]{1F77B4}{(-0.048)} & 2.708 \color[HTML]{1F77B4}{(-0.027)} \\
        S to C & 4.360 \color[HTML]{1F77B4}{(-0.038)} & 2.717 \color[HTML]{1F77B4}{(-0.018)} \\
        \hline
        S (baseline) & 4.543 & 2.829 \\
        M to S & 4.512 \color[HTML]{1F77B4}{(-0.031)} & 2.804 \color[HTML]{1F77B4}{(-0.025)} \\
        C to S & 4.529 \color[HTML]{1F77B4}{(-0.014)} & 2.811 \color[HTML]{1F77B4}{(-0.018)} \\
        S to S & 4.525 \color[HTML]{1F77B4}{(-0.018)} & 2.806 \color[HTML]{1F77B4}{(-0.023)} \\
        \Xhline{2\arrayrulewidth}
    \end{tabular}
    }
    \caption{\label{tab:WFLW}  
        Evaluation table on WFLW dataset.
    }
\end{table}

\begin{table}[tp]
    \centering%
    \tiny
    \renewcommand{\arraystretch}{0.85}
    \resizebox{0.35\textwidth}{!}{
    \begin{tabular}{ l | l  l }
        \Xhline{2\arrayrulewidth}
        \textbf{Structure} & intra-C variance \(\downarrow\) & TAR@FAR=0.1\% \(\uparrow\) \\
        \hline
        \hline
        M (baseline) & 0.6310 & 0.3678 \\
        M to M & 0.3186 \color[HTML]{1F77B4}{(-0.312)} & 0.4938 \color[HTML]{1F77B4}{(+12.60\%)} \\
        C to M & 0.3874 \color[HTML]{1F77B4}{(-0.243)} & 0.4959 \color[HTML]{1F77B4}{(+12.81\%)} \\
        S to M & 0.3158 \color[HTML]{1F77B4}{(-0.315)} & 0.5248 \color[HTML]{1F77B4}{(+15.70\%)} \\
        \hline
        C (baseline) & 0.1714 & 0.7645 \\
        M to C & 0.1632 \color[HTML]{1F77B4}{(-0.008)} & 0.7893 \color[HTML]{1F77B4}{(+2.48\%)} \\
        C to C & 0.1668 \color[HTML]{1F77B4}{(-0.004)} & 0.7934 \color[HTML]{1F77B4}{(+2.89\%)} \\
        S to C & 0.1738 \color[HTML]{D62728}{(+0.002)} & 0.7686 \color[HTML]{1F77B4}{(+0.41\%)} \\
        \hline
        S (baseline) & 0.4580 & 0.6488 \\
        M to S & 0.3493 \color[HTML]{1F77B4}{(-0.108)} & 0.6921 \color[HTML]{1F77B4}{(+4.33\%)} \\
        C to S & 0.4518 \color[HTML]{1F77B4}{(-0.006)} & 0.6302 \color[HTML]{D62728}{(-1.86\%)} \\
        S to S & 0.5966 \color[HTML]{D62728}{(+0.138)} & 0.6507 \color[HTML]{1F77B4}{(+0.82\%)} \\
        \Xhline{2\arrayrulewidth}
    \end{tabular}
    }
    \caption{\label{tab:CFP-dataset}  
        Evaluation table on CFP-dataset.
    }
\end{table}

For face recognition evaluation, we employed accuracy, F1-score, intra-class variance, and true acceptance rate (TAR), a metric commonly employed in biometric recognition. Validation on the CFP dataset, which contains a high proportion of challenging examples, was performed at a false acceptance rate (FAR) of 0.1\%. Identity classification demonstrated performance gains in both accuracy and F1-score within CNN architectures. The Swin architecture exhibited a mixed pattern of performance gains and losses, which we attribute to the weak inductive biases associated with ViT when applied to recognition tasks. Notably, MobileOne with KT-Adapter exhibited dramatic performance improvements, with intra-class variance reduced by over half and TAR gains over 10\%. While "S to C" and "S to S" exhibited minor decreases in intra-class variance, TAR exhibited minor increases. This observation suggests improved discriminative power and feature representation within the inter-class space.

To assess face landmark detection performance on the WFLW dataset, normalized mean error (NME) and mean absolute error (MAE) on 98 landmark points were adopted as evaluation metrics. The NME was normalized by the inter-ocular distance, defined as the Euclidean distance between the lateral canthi of the left and right eyes. The MAE was scaled relative to the input size of 224. Although significant performance enhancements were not universally demonstrated in Tab.~\ref{tab:WFLW}, all metrics exhibited a statistically significant improvement with the KT-Adapter methodology, with the exception of NME of "S to M" with +0.004. Notably, the adoption of CNN-based task priors consistently yielded more pronounced performance compared to ViT-based priors.

\begin{table}[tp]
    \centering%
    \tiny
    \renewcommand{\arraystretch}{0.85}
    \resizebox{0.35\textwidth}{!}{%
    \begin{tabular}{ l | l  l }
        \Xhline{2\arrayrulewidth}
        \textbf{Structure} & CS@5 \(\uparrow\) & MAE \(\downarrow\) \\
        \hline
        \hline
        M (baseline) & 66.79 & 4.568 \\
        M to M & 67.29 \color[HTML]{1F77B4}{(+0.50\%)} & 4.548 \color[HTML]{1F77B4}{(-0.020)} \\
        C to M & 66.85 \color[HTML]{1F77B4}{(+0.06\%)} & 4.549 \color[HTML]{1F77B4}{(-0.019)} \\
        S to M & 67.24 \color[HTML]{1F77B4}{(+0.45\%)} & 4.554 \color[HTML]{1F77B4}{(-0.014)} \\
        \hline
        C (baseline) & 68.32 & 4.316 \\
        M to C & 68.66 \color[HTML]{1F77B4}{(+0.34\%)} & 4.285 \color[HTML]{1F77B4}{(-0.031)} \\
        C to C & 69.10 \color[HTML]{1F77B4}{(+0.78\%)} & 4.288 \color[HTML]{1F77B4}{(-0.028)} \\
        S to C & 69.30 \color[HTML]{1F77B4}{(+0.98\%)} & 4.277 \color[HTML]{1F77B4}{(-0.039)} \\
        \hline
        S (baseline) & 67.49 & 4.412 \\
        M to S & 67.53 \color[HTML]{1F77B4}{(+0.04\%)} & 4.402 \color[HTML]{1F77B4}{(-0.010)} \\
        C to S & 67.60 \color[HTML]{1F77B4}{(+0.11\%)} & 4.400 \color[HTML]{1F77B4}{(-0.012)} \\
        S to S & 67.49 (\(\pm\)0\%) & 4.402 \color[HTML]{1F77B4}{(-0.010)} \\
        \Xhline{2\arrayrulewidth}
    \end{tabular}
    }
    \caption{\label{tab:UTK-face}  
        Evaluation table on UTK-face dataset. 
    }
\end{table}

\begin{table}[tp]
    \centering%
    \tiny
    \renewcommand{\arraystretch}{0.85}
    \resizebox{0.35\textwidth}{!}{
    \begin{tabular}{ l | l  l }
        \Xhline{2\arrayrulewidth}
        \textbf{Structure} & Accuracy \(\uparrow\) & F1-score \(\uparrow\) \\
        \hline
        \hline
        M (baseline) & 72.61 & 70.23 \\
        M to M & 73.09 \color[HTML]{1F77B4}{(+0.48\%)} & 70.52 \color[HTML]{1F77B4}{(+0.29\%)} \\
        C to M & 72.99 \color[HTML]{1F77B4}{(+0.38\%)} & 70.40 \color[HTML]{1F77B4}{(+0.17\%)} \\
        S to M & 73.23 \color[HTML]{1F77B4}{(+0.62\%)} & 70.63 \color[HTML]{1F77B4}{(+0.40\%)} \\
        \hline
        C (baseline) & 74.67 & 72.33 \\
        M to C & 74.34 \color[HTML]{D62728}{(-0.33\%)} & 72.04 \color[HTML]{D62728}{(-0.29\%)} \\
        C to C & 74.65 \color[HTML]{D62728}{(-0.02\%)} & 72.35 \color[HTML]{1F77B4}{(+0.02\%)} \\
        S to C & 74.72 \color[HTML]{1F77B4}{(+0.05\%)} & 72.45 \color[HTML]{1F77B4}{(+0.12\%)} \\
        \hline
        S (baseline) & 73.90 & 71.76 \\
        M to S & 75.13 \color[HTML]{1F77B4}{(+1.23\%)} & 72.76 \color[HTML]{1F77B4}{(+1.00\%)} \\
        C to S & 75.20 \color[HTML]{1F77B4}{(+1.30\%)} & 72.77 \color[HTML]{1F77B4}{(+1.01\%)} \\
        S to S & 75.01 \color[HTML]{1F77B4}{(+1.11\%)} & 72.63 \color[HTML]{1F77B4}{(+0.87\%)} \\
        \Xhline{2\arrayrulewidth}
    \end{tabular}
    }
    \caption{\label{tab:Affect-Net-HQ}  
        Evaluation table on Affect-Net-HQ dataset.
    }
\end{table}

For the evaluation of age estimation performance, MAE and cumulative score (CS) metrics were employed on the UTK-face dataset. The MAE was computed as the absolute deviation between predicted and ground truth ages, spanning the range of 1 to 116 years. The CS metric was determined as the percentage of predictions falling within \(\pm\)2 years, a tolerance level of 5. Consistent with observations in landmark detection, MAE exhibited marginal performance enhancements, lacking substantial improvements in Tab.~\ref{tab:CFP-dataset}. However, the CS metric demonstrated statistically significant performance gains. Notably, the adaptation of task priors among high-capacity models within the ConvNeXt architecture and Swin Transformer yielded superior performance improvements, "C to C" with +0.78\% and "S to C" +0.98\%, compared to alternative architectures, which achieves substantial performance gains relative to KT-Adapter's minimal computational overhead.

    

\begin{table*}[htp]
    \centering%
    \tiny
    \renewcommand{\arraystretch}{0.85}
    \resizebox{1.0 \textwidth}{!}{
    \begin{tabular}{ c | l l | l l | l l | l l}
        \Xhline{2\arrayrulewidth}
         & \multicolumn{2}{c}{\textbf{WFLW}} & \multicolumn{2}{c}{\textbf{UTK-face}} & \multicolumn{2}{c}{\textbf{Affect-Net-HQ}} & \multicolumn{2}{c}{\textbf{CFP-dataset}} \\
        \hline
        \textbf{Structure} & NME \(\downarrow\) & MAE \(\downarrow\) & CS@5 \(\uparrow\) & MAE \(\downarrow\) & Accuracy \(\uparrow\) & F1-score \(\uparrow\) & intra-C-var \(\downarrow\) & TAR@FAR=0.1\% \(\uparrow\) \\
        \hline
        \hline
        M (baseline) & 5.016 & 3.104 & 66.79 & 4.568 & 72.61 & 70.23 & 0.6310 & 36.78 \\
        1 encoder & 4.985 \color[HTML]{1F77B4}{(-0.030)} & 3.084 \color[HTML]{1F77B4}{(-0.020)} & 66.28 \color[HTML]{1F77B4}{(+0.49\%)} & 4.548 \color[HTML]{1F77B4}{(-0.019)} & 73.08 \color[HTML]{1F77B4}{(+0.47\%)} & 70.50 \color[HTML]{1F77B4}{(+0.28\%)} & 0.4627 \color[HTML]{1F77B4}{(-0.168)} & 49.44 \color[HTML]{1F77B4}{(+12.67\%)}\\
        2 encoders & 4.976 \color[HTML]{1F77B4}{(-0.039)} & 3.080 \color[HTML]{1F77B4}{(-0.024)} & 66.85 \color[HTML]{1F77B4}{(+0.05\%)} & 4.550 \color[HTML]{1F77B4}{(-0.017)} & 72.98 \color[HTML]{1F77B4}{(+0.37\%)} & 70.39 \color[HTML]{1F77B4}{(+0.17\%)} & 0.6090 \color[HTML]{1F77B4}{(-0.021)} & 49.13 \color[HTML]{1F77B4}{(+12.36\%)}\\
        3 encoders(proposed) & 4.976 \color[HTML]{1F77B4}{(-0.040)} & 3.080 \color[HTML]{1F77B4}{(-0.024)} & 67.29 \color[HTML]{1F77B4}{(+0.50\%)} & 4.548 \color[HTML]{1F77B4}{(-0.020)} & 73.09 \color[HTML]{1F77B4}{(+0.48\%)} & 70.63 \color[HTML]{1F77B4}{(+0.40\%)} & 0.3186 \color[HTML]{1F77B4}{(-0.312)} & 52.48 \color[HTML]{1F77B4}{(+15.70\%)}\\
        \hline
        C (baseline) & 4.398 & 2.735 & 68.32 & 4.316 & 74.67 & 72.33 & 0.1714 & 76.45 \\
        1 encoder & 4.361 \color[HTML]{1F77B4}{(-0.036)} & 2.713 \color[HTML]{1F77B4}{(-0.021)} & 68.65 \color[HTML]{1F77B4}{(+0.32\%)} & 4.285 \color[HTML]{1F77B4}{(-0.030)} & 74.34 \color[HTML]{D62728}{(-0.33\%)} & 72.04 \color[HTML]{D62728}{(-0.29\%)} & 0.1633 \color[HTML]{1F77B4}{(-0.008)} & 78.91 \color[HTML]{1F77B4}{(+2.47\%)}\\
        2 encoders & 4.352 \color[HTML]{1F77B4}{(-0.045)} & 2.709 \color[HTML]{1F77B4}{(-0.025)} & 69.05 \color[HTML]{1F77B4}{(+0.72\%)} & 4.289 \color[HTML]{1F77B4}{(-0.026)} & 74.65 \color[HTML]{D62728}{(-0.02\%)} & 72.34 \color[HTML]{1F77B4}{(+0.02\%)} & 0.1672 \color[HTML]{1F77B4}{(-0.004)} & 79.14 \color[HTML]{1F77B4}{(+2.69\%)}\\
        3 encoders(proposed) & 4.350 \color[HTML]{1F77B4}{(-0.048)} & 2.708 \color[HTML]{1F77B4}{(-0.027)} & 69.30 \color[HTML]{1F77B4}{(+0.98\%)} & 4.277 \color[HTML]{1F77B4}{(-0.039)} & 74.72 \color[HTML]{1F77B4}{(+0.05\%)} & 72.45 \color[HTML]{1F77B4}{(+0.12\%)} & 0.1632 \color[HTML]{1F77B4}{(-0.008)} & 79.34 \color[HTML]{1F77B4}{(+2.89\%)}\\
        \hline
        S (baseline) & 4.543 & 2.829 & 67.49 & 4.412 & 73.90 & 71.76 & 0.4580 & 64.88 \\
        1 encoder & 4.513 \color[HTML]{1F77B4}{(-0.029)} & 2.804 \color[HTML]{1F77B4}{(-0.024)} & 67.48 \color[HTML]{D62728}{(-0.10\%)} & 4.402 \color[HTML]{1F77B4}{(-0.010)} & 75.09 \color[HTML]{1F77B4}{(+1.19\%)} & 72.74 \color[HTML]{1F77B4}{(+0.99\%)} & 0.3497 \color[HTML]{1F77B4}{(-0.108)} & 69.11 \color[HTML]{1F77B4}{(+4.23\%)}\\
        2 encoders & 4.530 \color[HTML]{1F77B4}{(-0.012)} & 2.812 \color[HTML]{1F77B4}{(-0.016)} & 67.60 \color[HTML]{1F77B4}{(+0.10\%)} & 4.401 \color[HTML]{1F77B4}{(-0.011)} & 75.15 \color[HTML]{1F77B4}{(+1.26\%)} & 72.71 \color[HTML]{1F77B4}{(+0.96\%)} & 0.4523 \color[HTML]{1F77B4}{(-0.005)} & 63.12 \color[HTML]{D62728}{(-1.75\%)}\\
        3 encoders(proposed) & 4.512 \color[HTML]{1F77B4}{(-0.031)} & 2.804 \color[HTML]{1F77B4}{(-0.025)} & 67.60 \color[HTML]{1F77B4}{(+0.11\%)} & 4.400 \color[HTML]{1F77B4}{(-0.012)} & 75.20 \color[HTML]{1F77B4}{(+1.30\%)} & 72.77 \color[HTML]{1F77B4}{(+1.01\%)} & 0.3493 \color[HTML]{1F77B4}{(-0.108)} & 69.21 \color[HTML]{1F77B4}{(+4.33\%)}\\
        \Xhline{2\arrayrulewidth}
    \end{tabular}
    }
    \caption{\label{tab:ablation-N}  
        Ablation study on parameter \(N\), the number of encoders. We reported the results using the best-performing encoder among those sharing the same architecture type and encoder count. For example, in the face recognition task, the best performance was achieved when using encoders on age estimation and face landmark detection. We observed that increasing the number of prior encoders generally led to a linear improvement in performance.
    }
\end{table*}

\begin{table}[tp]
    \centering%
    \scriptsize
    \resizebox{0.45\textwidth}{!}{
    \begin{tabular}{ c c c | l  l | l  l }
        \Xhline{2\arrayrulewidth}
        \multicolumn{3}{c}{\textbf{Structure}} & \multicolumn{2}{c}{\textbf{Affect-Net-HQ}} & \multicolumn{2}{c}{\textbf{UTK-face}} \\
        \hline
        DM & CM & PE & Acc \(\uparrow\) & F1 \(\uparrow\) & CS@5 \(\uparrow\) & MAE \(\downarrow\) \\
        \hline
        \hline
        \ding{55} & \ding{55} & \ding{55} &  0.7022 & 0.6677 &  62.50 & 4.771 \\
        \ding{51} & \ding{55} & \ding{55} &  0.7090 & 0.6998 &  66.00 & 4.601 \\
        \ding{51} & \ding{55} & \ding{51} &  0.7095 & 0.7002 &  65.89 & 4.588 \\
        \ding{55} & \ding{51} & \ding{51} &  0.7152 & 0.7020 &  66.51 & 4.599 \\
        \ding{55} & \ding{51} & \ding{51} &  0.7199 & 0.7022 &  66.51 & 4.598 \\
        \ding{51} & \ding{51} & \ding{51} &  0.7283 & 0.7041 &  66.87 & 4.550 \\
        \ding{51} & \ding{51} & \ding{55} &  0.7299 & 0.7040 &  66.85 & 4.549 \\
        \Xhline{2\arrayrulewidth}
    \end{tabular}
    }
    \caption{\label{tab:ablation-param}  
        Ablation study on proposed modules.
    }
\end{table}

For the classification task of face emotion recognition, we employed standard evaluation metrics, accuracy and F1-score. As shown in Tab.~\ref{tab:Affect-Net-HQ}, all methodologies, with the exception of ConvNeXt, exhibited performance enhancements. We posit that this phenomenon is attributable to the inherent high capacity of "C" for the emotion recognition task relative to the other task priors. Notably, despite "S" demonstrating significantly lower accuracy compared to "C" (0.7390 and 0.7176 vs. 0.7467 and 0.7233), the application of KT-Adapter in the "C to S" outperformed with superior performance (0.7520 and 0.7277). This outcome is particularly compelling, as the observed performance gains in the "S to C", where the task priors are reversed to "C to S", suggest a synergistic relationship between ConvNeXt and Swin for face emotion recognition.

\subsection{Ablation Studies}
An ablation study was conducted to assess the effects of the number of encoders N in Tab.~\ref{tab:ablation-N}, the use of drop mask "DM", canonical mask "CM", and the positional encoding "PE" in Tab.~\ref{tab:ablation-param}. Evaluation metrics included accuracy and F1-score for facial emotion recognition, and CS@5 and MAE for age estimation. The use of the drop mask alone yielded improvements compared to the baseline model. Applying the canonical mask alone also yielded performance gains. Notably, a synergistic effect was observed when both the drop mask and canonical mask were employed simultaneously. As discussed in Sec.~\ref{sec:decoder-and-inference}, this performance improvement indicates that structural misalignment between the training and inference phases plays a significant role in KT-Adapter. Interestingly, incorporating positional encoding into the prior tokens had only a minimal impact on performance. Considering the computational efficiency benefits of drop mask and canonical mask in training, we exclude positional encoding to reduce computational overhead.

\section{Conclusion}
The KT-Adapter exhibited performance gains across tasks and evaluation metrics, while maintaining a demonstrably minimal computational overhead, with particularly notable improvements observed at the feature representation level. However, a salient limitation inherent in knowledge token methodologies is their susceptibility to overfitting datasets. The training phase, which relies on task priors and presents each encoder with canonical input, potentially impedes the learning of multi-modal or multi-input modalities. We posit that this research serves as a cornerstone towards the effective exploitation of task priors and encourage future investigations to broaden its applicability beyond the domain of facial analysis, extending to more generalized domains.